\documentclass{article}
\usepackage{spconf,amsmath,amsfonts,graphicx}
\usepackage{threeparttable}
\usepackage{subfiles}
\usepackage{multirow}
\usepackage{stfloats}
\usepackage{url}
\usepackage{colortbl}
\usepackage{xcolor}
\usepackage{soul}
\usepackage[hidelinks]{hyperref}
\usepackage{booktabs}
\usepackage{array}

\usepackage{caption}
\usepackage{float}
\restylefloat{table}

\DeclareMathOperator{\Enc}{Enc}
\DeclareMathOperator{\CTC}{CTC}
\DeclareMathOperator{\CE}{CE}
\DeclareMathOperator{\Dec}{Pred}
\DeclareMathOperator{\Joint}{Joint}

\DeclareMathOperator{\BiLSTM}{BiLSTM}
\DeclareMathOperator{\softmax}{softmax}

\newcommand{\nobias}{\texttt{\textless no\_bias\textgreater}}

\title{Improving ASR Contextual Biasing with Guided Attention}
\name{
Jiyang Tang$^{2*}$\thanks{$^*$Work done during an internship at ASAPP.},
Kwangyoun Kim$^1$,
Suwon Shon$^1$,
Felix Wu$^1$,
Prashant Sridhar$^1$,
Shinji Watanabe$^2$
}

\address{
  $^1$ASAPP Inc., New York, NY, USA\\
  $^2$Carnegie Mellon University, Pittsburgh, PA, USA}

\begin{document}
\ninept

\setlength{\abovedisplayskip}{4pt}
\setlength{\belowdisplayskip}{4pt}

\maketitle
 
\begin{abstract}

In this paper, we propose a Guided Attention (GA) auxiliary training loss, which improves the effectiveness and robustness of automatic speech recognition (ASR) contextual biasing without introducing additional parameters.
A common challenge in previous literature is that the word error rate (WER) reduction brought by contextual biasing diminishes as the number of bias phrases increases.
To address this challenge, we employ a GA loss as an additional training objective besides the Transducer loss. 
The proposed GA loss aims to teach the cross attention how to align bias phrases with text tokens or audio frames.
Compared to studies with similar motivations, the proposed loss operates directly on the cross attention weights and is easier to implement.
Through extensive experiments based on Conformer Transducer with Contextual Adapter, we demonstrate that the proposed method not only leads to a lower WER but also retains its effectiveness as the number of bias phrases increases.
Specifically, the GA loss decreases the WER of rare vocabularies by up to 19.2\% on LibriSpeech compared to the contextual biasing baseline, and up to 49.3\% compared to a vanilla Transducer.

\end{abstract}

\section{Introduction}\label{sec:intro}

End-to-end automatic speech recognition (ASR) architectures have gained substantial interest among research communities and in industrial applications in recent years.
Models based on connectionist temporal classification
(CTC)~\cite{CTC}, recurrent neural network transducer (RNN-T)~\cite{RNNT,asr_dnn}, and attention-based encoder-decoder (AED)~\cite{asr_att,LAS} have become the foundation of state-of-the-art ASR systems.
Although these models can achieve impressively low word error rates (WER), recognizing uncommon words such as proper nouns and acronyms still remains a challenging problem.
In particular, practical applications such as voice assistants and automated customer services may not operate properly due to insufficient information.
Therefore, supplying ASR models with prior knowledge about possible occurrences of rare vocabularies, a process called \textit{contextual biasing}, is a cheap and effective way to improve the recognition accuracy of these uncommon phrases.

Several lines of research have been proposed to perform contextual biasing.
The first category of research relies on the shallow fusion of external language model and on-the-fly rescoring during decoding to achieve contextual biasing~\cite{bruguier16_interspeech,DucLe21_slt,Le21,zhao19d_interspeech,Huang2020ClassLA}.
The second category involves the integration of the context in ASR models.
The early work includes~\cite{DeepContext,Jain2020,Chang21}.
Later, Contextual Adapter is proposed in~\cite{ContextualAdapters} and becomes one of the most popular methods for integrated contextual biasing with many derivative studies~\cite{MultilingualCA,AdaptiveCB,GatedCA,CACPP,Munkhdalai22,Fu23,Tong23}.
A major challenge among these studies is that the model struggles to select the correct bias phrases if the number of phrases given to the model is large, reducing the effectiveness of contextual biasing.
A common method to address this challenge is to add new training objectives.
\cite{CACPP} uses a contextual phrase prediction network to predict the text occurrence of bias phrases.
\cite{AdaptiveCB} adds an additional module to detect if a bias phrase occurred in the utterance.
\cite{MultilingualCA} and \cite{CopyNE} forces the model to learn the alignment between bias phrases and text tokens or audio frames through auxiliary losses calculated from cross-attention scores.
A similar alignment method is adopted to Spoken Language Understanding.
For example, RNN-T or cross-entropy loss is used for aligning slot tags and text transcription to enhance the effectiveness of a multi-task model~\cite{Fu2022MultiTaskRW}.

In this paper, we address the challenge by providing explicit guidance to deep contextual biasing models.
More specifically, our research contributions include:
\begin{enumerate}
    \item We propose a simple and effective auxiliary training loss based on CTC that operates directly on the cross attention weights, called Guided Attention (GA) CTC loss.
    \item We apply the proposed loss to Conformer Transducer~\cite{RNNT} with Contextual Adapter.
    Through extensive experiments, we demonstrate that the proposed method not only leads to a lower WER but also retains the effectiveness of contextual biasing as the number of bias phrases increases.
    \item We compare our approach against an alternative formulation, showing that the proposed loss achieves the same effects but is easier to implement in practice.
\end{enumerate}

\section{Background}

In this section, we introduce the Transducer architecture~\cite{RNNT,asr_dnn} and how contextual biasing is enabled using Contextual Adapter~\cite{ContextualAdapters}.

\subsection{Neural-Transducer}

A Neural-Transducer network~\cite{RNNT,asr_dnn} usually consists of an encoder network, a prediction network, and a joint network.
The encoder network~$\Enc(\cdot)$ processes a sequence of acoustic feature vectors, $\mathbf{x} = (\mathbf{x}_1,\mathbf{x}_2,\dots,\mathbf{x}_T)$, and outputs a high-level representation $\mathbf{h}_t^{\Enc}$ at time $t$.
The prediction network~$\Dec(\cdot)$ produces a vector $\mathbf{h}_{u+1}^{\Dec}$ at $(u+1)$-th step based on the previous text sequence $\mathbf{y}_{1:u} = (\mathbf{y}_1,\mathbf{y}_2,\dots,\mathbf{y}_u)$ in an auto-regressive manner.
The joint network $\Joint(\cdot)$ merges the output of these two networks using a dense network and predicts the output token $P(\mathbf{y}_{u+1}|\mathbf{x}_t,\mathbf{y}_{1:u})$ in the form of posteriors over vocabulary $\mathcal{V}$:
\begin{align}
    \mathbf{h}_t^{\Enc} &= \Enc(\mathbf{x}_t) \\
    \mathbf{h}_{u+1}^{\Dec} &= \Dec(\mathbf{y}_u,\mathbf{h}_{u}^{\Dec}) \\
    P(\mathbf{y}_{u+1}|\mathbf{x}_t,\mathbf{y}_{1:u}) &= \Joint(\mathbf{h}_t^{\Enc}, \mathbf{h}_{u+1}^{\Dec}) \label{eq:joint_network}
\end{align}


\subsection{Contextual Adapter}
\label{sec:adapter}

Given a \textit{bias phrase list} $\mathbf{Q} = (\mathbf{q}_1,\mathbf{q}_2,\dots,\mathbf{q}_S)$, which is a set of bias phrases $\mathbf{q}_s\in \mathbb{Z}_{+}^{K_s}$ each containing $K_s$ text tokens, a contextual biasing module provides context information that helps the ASR model recognize phrases on that list more accurately from the audio input. For instance, when making a call using a voice assistant on a smart-device, it can utilize names from the contact list, or when requesting to play music, song titles or artist names can be added to the list as the target bias phrases. As a result, contextual biasing enables the model to recognize the phrases in the list more precisely.

\begin{figure}[ht]
\centering
\includegraphics[width=\linewidth]{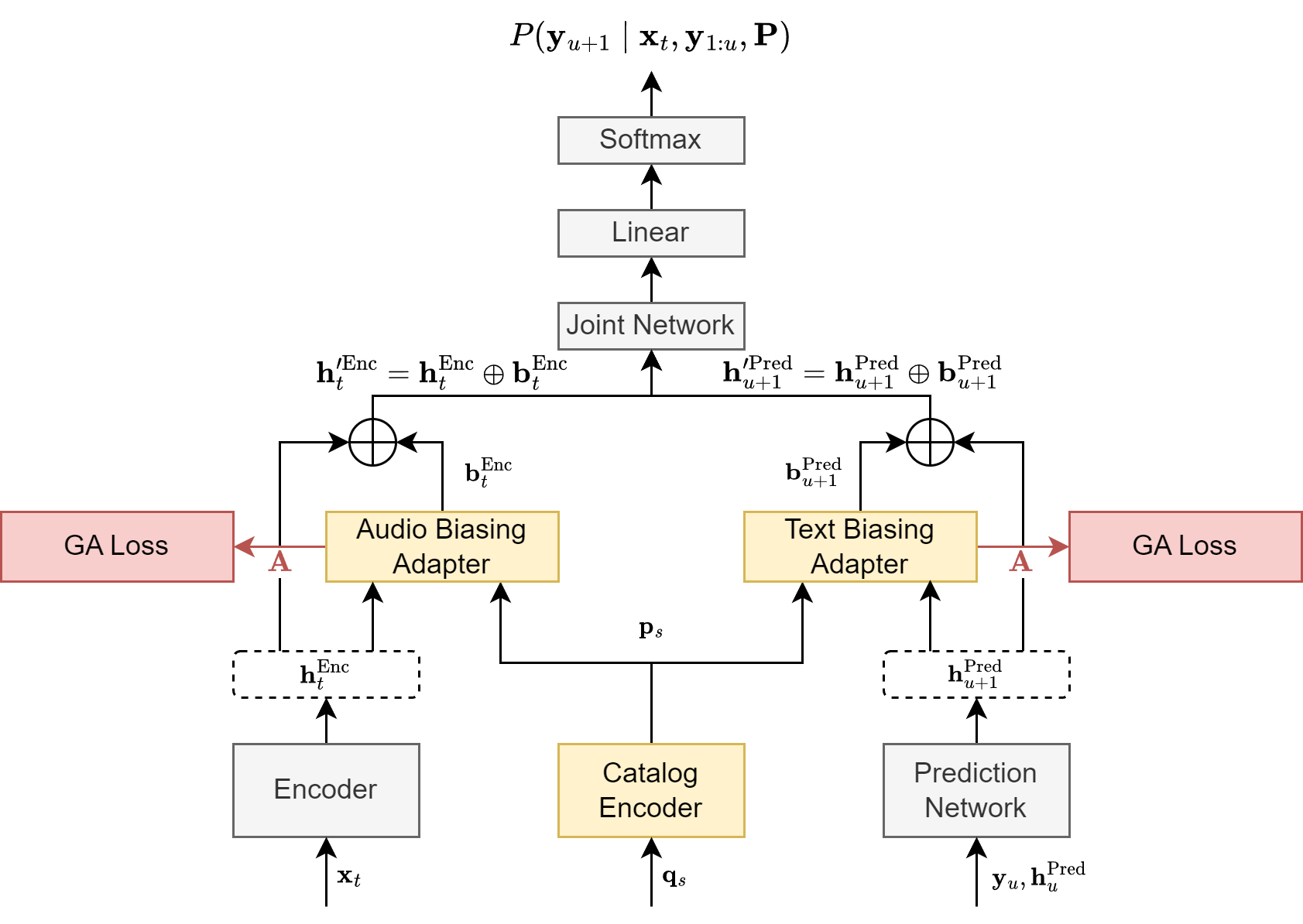}
\caption{
The architecture of Transducer with Contextual Adapter and the proposed Guided Attention (GA) loss.
During training, gray modules are frozen, while only the yellow modules are included in back-propagation.
The GA loss, highlighted in red, is newly applied to both the audio and the text biasing adapters as auxiliary losses.
}
\label{fig:arch}
\vspace{-4mm}
\end{figure}

Contextual Adapter~\cite{ContextualAdapters} is one of the most successful contextual biasing methods among recent studies.
Based on the Transducer architecture, it introduces two additional modules: the catalog encoder and the biasing adapter, as shown in Figure~\ref{fig:arch}.
The catalog encoder condenses every bias phrase $\mathbf{q}_s$ into a one-dimensional \textit{phrase embedding} $\mathbf{p}_s\in\mathbb{R}^{E}$ by taking the last hidden state of a bidirectional LSTM model~\cite{ContextualAdapters}, where $E$ is the embedding size.
By doing this, we convert a list of variable-length text sequences into a list of fixed-size vector representations:
\[
    \mathbf{p}_s = \BiLSTM(\mathbf{q}_s)
\]
Given this list of phrase embeddings $\mathbf{P} = (\mathbf{p}_1,\mathbf{p}_2,\dots,\mathbf{p}_S)$ with size $S$, the biasing layer uses multi-head attention (MHA) mechanism to align the bias phrases with the text sequence or with the audio frames.
Specifically, the biasing layer is inserted between the encoder network and the joint network, and also between the prediction network and the joint network, as shown in Figure~\ref{fig:arch}.
Looking at the audio biasing adapter first,
we compute the attention scores $\mathbf{A}_{t}$  at time $t$ using $\mathbf{h}_t^{\Enc}$ for the attention query and $\mathbf{P}$ for the key:
\begin{equation}
    \mathbf{A}_{t} = \softmax \left( \frac{\mathbf{h}_t^{\Enc}\mathbf{W}^Q(\mathbf{P}\mathbf{W}^K)^T}{\sqrt{d}} \right)
    \label{eq:att_score}
\end{equation}
where $\mathbf{W}^K, \mathbf{W}^Q \in \mathbb{R}^{E \times E}$ are the projection matrices for key and query, respectively, and $d$ is the hidden dimension.
Here we consider the case of single-head attention for simplicity, but the equations also apply to MHA.
Then we calculate the weighted sum of attention values to obtain the encoder \textit{biasing vector} $\mathbf{b}_t^{\Enc}$:
\begin{equation}
    \mathbf{b}_t^{\Enc} = \mathbf{A}_{t} \mathbf{P} \mathbf{W}^V
\end{equation}
where $\mathbf{W}^V \in \mathbb{R}^{E \times E}$ is the projection matrix for value.
After that, we obtain the context-aware encoder representation $\mathbf{h'}_t^{\Enc}$ by computing the element-wise addition between $\mathbf{h}_t^{\Enc}$ and $\mathbf{b}_t^{\Enc}$:
\begin{equation}
    \mathbf{h'}_t^{\Enc} = \mathbf{h}_t^{\Enc} \oplus \mathbf{b}_t^{\Enc}
\end{equation}
Meanwhile, we perform the same process for the prediction network to get $\mathbf{h'}_{u}^{\Dec}$ from $\mathbf{h}_{u}^{\Dec}$.
Finally, we feed the joint network with context-aware representations to produce context-aware token predictions.
So Equation~\ref{eq:joint_network} becomes:
\begin{equation}
P(\mathbf{y}_{u+1}|\mathbf{x}_t,\mathbf{y}_{1:u},\mathbf{P}) = \Joint(\mathbf{h'}_t^{\Enc}, \mathbf{h'}_{u+1}^{\Dec})
\end{equation}
It is a common practice to add a special \nobias phrase to the top of the bias list.
This allows the adapter to properly handle tokens or audio frames that do not correspond to any bias phrase~\cite{ContextualAdapters}.

\section{Guided Attention Loss}\label{sec:proposed}

As mentioned in Section~\ref{sec:intro}, providing additional training objectives can improve the effectiveness of Contextual Adapter.
\cite{MultilingualCA} proposes applying a cross entropy based auxiliary training loss directly to the attention score matrix to teach the biasing adapters how to better align bias phrases with the audio frames.
And this method is proven to be crucial for training Contextual Adapter modules in low-resource scenarios.
However, their investigation of this method is limited to the CTC encoder architecture~\cite{MultilingualCA}.
In this section, we describe how to adapt this loss to the Neural-Transducer architecture, discuss its shortcomings, and propose a simple yet effective method based on CTC.
The adapted cross entropy loss and the CTC variant are referred to as \textbf{Guided Attention Cross Entropy (GA-CE)} loss and \textbf{Guided Attention CTC (GA-CTC)} loss respectively.

\subsection{GA-CE Loss}\label{sec:ga_ce}
Let the attention score matrices of the audio and the text biasing adapter be
$\mathbf{A^{\Enc}} \in \mathbb{R}^{T \times {(S+1)}}$
and $\mathbf{A^{\Dec}} \in \mathbb{R}^{U \times {(S+1)}}$,
where $T$ and $U$ are the lengths of audio frames and text tokens, respectively, and $S$ is the size of the bias list excluding \nobias.
For GA-CE, we define the ordered index sequence of bias phrases for the audio and the text biasing adapter, $C^{\Enc} = (c_1^{\Enc}, \dots, c_T^{\Enc})$ and $C^{\Dec} = (c_1^{\Dec}, \dots, c_U^{\Dec})$ respectively, where $c\in[0,S]$ denotes the index at which a phrase appears in the bias list with regard to each audio frame at time $t$ or text token at step $u$, and $c=0$ corresponds to \nobias.
Then we treat each element $\mathbf{A}_{ts}$ in $\mathbf{A^{\Enc}}$ or $\mathbf{A^{\Dec}}$ as the likelihood of observing phrase with index $s$ at time $t$ and calculate GA-CE loss as follows:
\begin{equation}
    \mathcal{L}_{\text{GA-CE}} = \CE(\mathbf{A}^{\Enc}, C^{\Enc}) + \CE(\mathbf{A}^{\Dec}, C^{\Dec})
\end{equation}
The overall objective function is the weighted sum of the Transducer loss and GA-CE loss with weight $\alpha$ being a hyper-parameter:
\begin{equation}
\mathcal{L} = \alpha \mathcal{L}_{\text{GA-CE}} + (1-\alpha) \mathcal{L}_{\text{Transducer}}
\label{eq:ga_ce}
\end{equation}

There are two differences between this adapted formulation and the original one in~\cite{MultilingualCA}.
Firstly, the former is applied to both the audio biasing adapter and the text biasing adapter (Figure~\ref{fig:arch}), while the architecture used in~\cite{MultilingualCA} involves only an encoder biasing adapter.
Secondly, in ~\cite{MultilingualCA}, cross entropy of audio frames or text tokens are weighted by the complement of the attention score of \nobias at the same time step.
By doing this, only the frames or tokens where bias phrases occurred have meaningful contribution to the loss.
Meanwhile, our method calculates unweighted cross entropy for all audio frames or text tokens so that equal importance is assigned to each step whether contextual biasing is enabled or not.

\subsection{GA-CTC Loss}\label{sec:ga_ctc}

One of the tricky parts of utilizing GA-CE loss is that $C^{\Enc}$ is difficult to obtain for audio frames, as it usually involves performing forced alignment using an ASR model.
Motivated by this, we propose a simpler alternative $\mathcal{L}_{\text{GA-CTC}}$ using CTC~\cite{CTC}.
We first obtain $C = (c_1, \dots, c_L)$ easily without any extra processing like forced alignment, where $c\in[1,S]$ without \nobias or consecutively repeated indices, and $L$ is the length of the bias list indices for a certain input.
Then we calculate GA-CTC loss using $C$ as the label:
\begin{align}
    \mathcal{L}_{\text{GA-CTC}} &= \CTC(\mathbf{A}^{\Enc}, C) + \CTC(\mathbf{A}^{\Dec}, C)
\end{align}
The overall objective function becomes:
\begin{equation}
\mathcal{L} = \alpha \mathcal{L}_{\text{GA-CTC}} + (1-\alpha) \mathcal{L}_{\text{Transducer}}
\label{eq:ga_ctc}
\end{equation}

It's crucial to understand that \nobias is the equivalent of \texttt{$<$blank$>$} symbol for CTC.
This allows the module to attend to \nobias at any given time due to the rules of CTC~\cite{CTC}.
Because of this, the model can enable or disable contextual biasing at the correct time if trained properly.

To apply GA-CE or GA-CTC loss to an MHA module, we use the average attention scores across all heads as $\mathbf{A}$.

\section{Experiments}\label{sec:exp}

\subsection{Data}
\label{sec:data}

Our experiments are conducted using LibriSpeech~\cite{librispeech}, which includes 960 hours of English speech.
This dataset has been used in several previous studies~\cite{Le21,AdaptiveCB,CACPP}.

We generate the bias phrase list for training and testing using a list of rare words from~\cite{Le21}.
This list is obtained by sorting all words in the training set by their frequencies and removing the 5000 most common ones~\cite{Le21}.
Besides that, we include \textbf{distractors}, randomly sampled entries from rare words that are not in the utterance, to test if the biasing adapter can identify which bias phrases are relevant.
During training, the bias list of an utterance is the union of rare words that appeared in all utterances in the same batch.
This method results in training bias lists with dynamic sizes and variable distractors while being easy to implement.
During testing, the bias list of each utterance contains rare words that occurred in that utterance plus a certain amount of distractors.

To evaluate the performance of contextual biasing, we follow previous literature to include three metrics: word error rate (\textbf{WER}), unbiased word error rate (\textbf{U-WER}), and biased word error rate (\textbf{B-WER}).
WER is the word error rate of all words in the test corpus, U-WER measures the WER of words that are not in the bias list, and B-WER equals the WER of words in the bias list~\cite{Le21,ContextualAdapters,GatedCA,CACPP}.
We expect a good contextual biasing model to have a lower B-WER without increasing its U-WER significantly.
Further, it should also have minimal B-WER degradation as we add more distractors.

\subsection{Experimental Setup}

We use a Transducer network~\cite{RNNT} as the foundation for contextual biasing.
The network includes a Conformer encoder~\cite{conformer} and an RNN prediction network.
All our Transducer experiments are initialized from a pre-trained checkpoint\footnote{\url{huggingface.co/espnet/chai_librispeech_asr_train_conformer-rnn_transducer_raw_en_bpe5000_sp}}
from ESPnet2's LibriSpeech recipe~\cite{espnet}.
The Conformer encoder has $12$ blocks, each having $2048$ hidden units and $8$ attention heads.
The prediction network contains a single LSTM layer with $512$ hidden units.
The joint network has one layer with a size of $640$.
For contextual biasing, the catalog encoder contains an embedding layer with size $512$, a single-layer BiLSTM with size $512$, and a linear projection layer.
The embedding layer is initialized with the same layer from the prediction network.
The biasing adapter contains an MHA module with an embedding size of $512$ and $4$ attention heads and two linear layers.
These linear layers project the input and output of the biasing adapter to match the embedding size and the subsequent layer size.
In our configuration, the input and output sizes are $512$ for both adapters.
While freezing other pre-trained Transducer parts, we only train the catalog encoder, the audio biasing adapter, and the text biasing adapters using the Adam optimizer to $30$ epochs with a learning rate of $0.001$.
If an auxiliary loss is enabled, we use $0.5$ for $\alpha$ in Equation~\ref{eq:ga_ce} or Equation~\ref{eq:ga_ctc}.
All experiments are conducted using ESPNet2~\cite{espnet}.

\noindent\textbf{Baseline:}
The baseline (\textbf{CA}) is a Conformer Transducer model with a Contextual Adapter according to Section~\ref{sec:adapter}.

\noindent\textbf{GA-CE:}
Based on the baseline, we add GA-CE loss to the training objective as described in Section~\ref{sec:ga_ce}.
We first generate the phrase index label for each token in the text sequence $C^{\Dec}$ programmatically.
This is used by the auxiliary loss for the prediction network adapter.
On the encoder side, we obtain the phrase index label for each audio frame $C^{\Enc}$ using CTC forced alignment~\cite{ctc_seg}.

\noindent\textbf{GA-CTC:}
We are interested to see if GA-CTC loss can lead to the same level of WER reduction compared to GA-CE loss.
Therefore, we perform another experiment by adding GA-CTC loss to the Contextual Adapter baseline.
Unlike GA-CE loss, the label sequences $C$ can be easily generated during data loading.

\noindent\textbf{Bias List Configurations:}
For each experiment mentioned above, we use different \textbf{number of distractors $N$} for evaluation, as described in Section~\ref{sec:data}.
In addition, we want to test whether the model can maintain its original performance when no bias phrases are given to the model.
Therefore, we perform additional inference runs using a bias list that contains \nobias as the only entry.
We use \textbf{empty bias list} to refer to this configuration.

\begin{table}[h]

\centering
\resizebox{0.95\linewidth}{!}{
\begin{tabular}{lcccc}
\toprule 
Model & dev-clean & dev-other & test-clean & test-other \\
\midrule

Transducer             & 1.7/10.0 & 5.4/20.7 & 1.9/10.2 & 4.9/22.0 \\

+ CA & 1.7/10.3 & 5.3/21.1 & 1.9/10.3 & 4.8/22.0 \\

\hspace{2mm}+ GA-CE  & 1.7/10.2 & 5.4/20.8 & 1.9/10.1 & 4.9/21.9 \\

\hspace{2mm}+ GA-CTC & 1.7/10.3 & 5.3/20.9 & 1.9/10.2 & 4.9/22.0 \\

\bottomrule
\end{tabular}
}

\caption{\label{tab:baselines}
WER breakdown of vanilla Transducer and contextual adapters given \textbf{an empty bias list}, following format: U-WER/B-WER.
CA stands for Contextual Adapter.
}
\end{table}

\begin{table*}[t!]

\centering\resizebox{1.0\linewidth}{!}{
\begin{threeparttable}
\begin{tabular}{l|ccc|ccc|ccc|ccc}
\toprule 
\multirow{2}{*}{Model}
& \multicolumn{3}{c|}{dev-clean}
& \multicolumn{3}{c|}{dev-other}
& \multicolumn{3}{c|}{test-clean}
& \multicolumn{3}{c}{test-other} \\

& $N=0$ & $N=100$ & $N=1000$
& $N=0$ & $N=100$ & $N=1000$
& $N=0$ & $N=100$ & $N=1000$
& $N=0$ & $N=100$ & $N=1000$ \\
\midrule

\multirow{2}{*}{Transducer$^\dagger$}
& 2.6 & - & -
& 6.8 & - & -
& 2.8 & - & -
& 6.6 & - & -
\\
& (1.7/10.0) & - & -
& (5.4/20.7) & - & -
& (1.9/10.2) & - & -
& (4.9/22.0) & - & -
\\

\multirow{2}{*}{+ CA}
& 1.9 & 2.0 & 2.4
& 5.6 & 5.9 & 6.5
& 2.1 & 2.3 & 2.6
& 5.3 & 5.6 & 6.3
\\
& (1.6/4.6) & (1.6/5.7) & (1.7/7.7)
& (5.1/10.0) & (5.3/12.5) & (5.5/16.3)
& (1.8/4.8) & (1.8/5.9) & (1.9/7.8)
& (4.6/11.4) & (4.8/13.4) & (5.1/17.6)
\\

\multirow{2}{*}{\hspace{2mm}+ GA-CE}
& 1.8 & 1.9 & 2.1
& 5.5 & 5.7 & 6.2
& 2.1 & 2.2 & 2.4
& 5.2 & 5.4 & 6.0
\\
& (1.5/4.0) & (1.5/4.6) & (1.6/6.1)
& (5.2/9.1) & (5.2/10.5) & (5.4/13.7)
& (1.8/4.4) & (1.8/5.1) & (1.9/6.4)
& (4.6/10.3) & (4.7/12.2) & (5.0/15.3)
\\

\multirow{2}{*}{\hspace{2mm}+ GA-CTC}
& 2.0 & 2.0 & 2.2
& 5.6 & 5.8 & 6.3
& 2.2 & 2.3 & 2.4
& 5.5 & 5.6 & 6.2
\\
& (1.6/5.2) & (1.6/5.5) & (1.8/6.4)
& (5.1/10.5) & (5.2/10.9) & (5.6/12.9)
& (1.8/5.3) & (1.9/5.7) & (2.0/6.3)
& (4.7/12.4) & (4.8/13.1) & (5.2/15.2)
\\

\bottomrule
\end{tabular}
\begin{tablenotes}
    \item[$^\dagger$] The value of $N$ is irrelevant since contextual biasing is not enabled for the pre-trained transducer.
\end{tablenotes}
\end{threeparttable}
}

\caption{\label{tab:transducer_adapter} The word error rate breakdown of the baselines and the proposed method following format: WER(U-WER/B-WER). Different numbers of distractors $N$ are tested. CA and GA stand for Contextual Adapter and Guided Attention. }
\end{table*}

\subsection{Results and Discussion}
\label{sec:results}

Before evaluating the effectiveness of contextual biasing, we first examine if the Contextual Adapter baseline negatively affects WER for vanilla ASR.
We decode models using the empty bias list and expect the WER of these experiments to be as close to that of the pre-trained Transducer as possible.
Table~\ref{tab:baselines} shows that the results match our expectation within the error margin for all contextual biasing experiments.
This implies that the adapters can correctly turn off contextual biasing if \nobias is being attended to.

\begin{figure}[htb]
\vspace{-6mm}
\centering
\includegraphics[width=\linewidth]{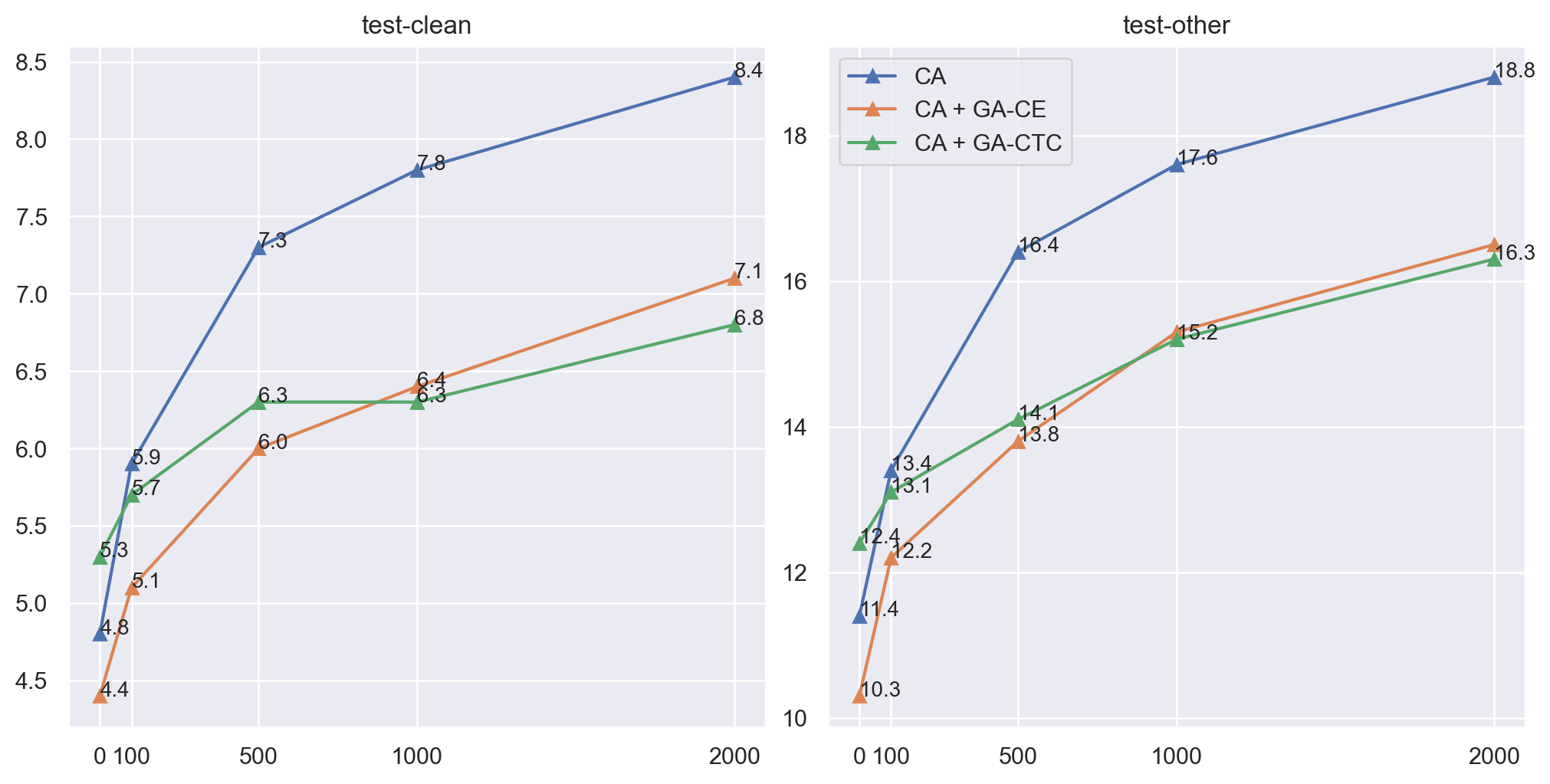}
\caption{
B-WER of Contextual Adapter (CA) baseline, CA with GA-CE loss and CA with GA-CTC loss, using different number of distractors $N\in\{0, 100, 500, 1000, 2000\}$.
}
\label{fig:bwer_vs_N}
\vspace{-4mm}
\end{figure}

Table~\ref{tab:transducer_adapter} and Figure~\ref{fig:bwer_vs_N} show the WER breakdown of baselines and the proposed method using different values for $N$.
Compared to the pre-trained transducer model, the Contextual Adapter baseline reduces B-WER by up to $54\%$ when $N=0$.
Meanwhile, it has minimal impact on U-WER.
This indicates that Contextual Adapter itself is effective at reducing B-WER without significantly increasing U-WER.
Thus, we focus on analyzing B-WER in this section.

As we increase $N$, it is evident that the baseline adapter becomes much less effective.
For example, the relative B-WER reduction brought by adapters is reduced from $48.2\%$ when $N=0$ to $20.0\%$ when $N=2000$ on \texttt{test-other}.
This suggests that Contextual Adapter trained without additional objectives is not robust against distractors.
As a result, this severely limits the practical use of contextual biasing as we often have bias lists that contain hundreds or thousands of bias phrases in production ASR systems.

Fortunately, GA loss can mitigate this issue.
According to Table~\ref{tab:transducer_adapter} and Figure~\ref{fig:bwer_vs_N}, both GA-CE and GA-CTC loss produce significantly lower B-WER when $N>0$ compared to without them.
In addition, the degree of increase in B-WER as $N$ increases is much lower.
This phenomenon is the most observable in Figure~\ref{fig:bwer_vs_N} when $N>100$.
As a result, we demonstrate that GA loss is crucial for maintaining B-WER reduction when there are multiple distractors.

Although both GA losses look promising, GA-CTC does not introduce B-WER reduction when $N=0$ compared to the CA baseline.
Moreover, GA-CTC loss is less effective than the GA-CE loss when $N<1000$, which indicates that relaxed alignment paths allowed by CTC are not as ideal for contextual biasing.
However, the performance gap between CTC and CE variants diminishes if $N\ge 1000$.
This implies that the performance bottleneck of adapters might be its ability to distinguish similar bias phrases, instead of the effectiveness of aligning bias phrases with the input sequences.
However, because GA-CTC loss is much easier to use in practice, it is more desirable for large bias phrase lists.

\begin{table}[ht]

\centering
\resizebox{1.0\linewidth}{!}{
\begin{tabular}{p{0.2\linewidth}|p{0.8\linewidth}}
\toprule
Model & Text \\
\midrule




\multirow{2}{*}{Ground Truth} & ... that would have destroyed the whole plant but \textbf{MARSHALL} never even thanked me \\
& Take him out \textbf{THORKEL} and let him taste your sword \\
\hline

\multirow{2}{*}{Transducer} & ... that would have destroyed the whole plant but MARTIAL never even thanked me \\
& Take him out TOKEL and let him taste your sword \\
\hline

\multirow{2}{*}{+ CA} & ... that would have destroyed the whole plant but \textbf{MARSHALL} never even thanked me \\
(N=0) & Take him out TORKEL and let him taste your sword \\
\hline

\multirow{2}{*}{+ CA} & ... that would have destroyed the whole plant but MARTIAL never even thanked me \\
(N=2000) & Take him out TOKEL and let him taste your sword \\
\hline

\multirow{2}{*}{\hspace{2mm}+ GA-CTC} & ... that would have destroyed the whole plant but \textbf{MARSHALL} never even thanked me \\
(N=2000) & Take him out \textbf{THORKEL} and let him taste your sword \\

\bottomrule
\end{tabular}
}

\caption{\label{tab:examples} Examples of recognition results using Contextual Adapter (CA) and the proposed Guided Attention Loss based on CTC (GA-CTC). $N$ denotes the size of distractors. The words in upper-case indicate the target words in the bias list, and words in bold are accurately recognized.
}
\end{table}

Table~\ref{tab:examples} demonstrates the effect of GA-CTC loss with two example utterances.
Comparing the vanilla Transducer and the CA baseline, it is evident that Contextual Adapter helps predict the uncommon words in the utterances correctly (``MARSHALL" and ``THORKEL'').
However, the CA baseline fails to align these two words with the corresponding bias list entries as we increase $N$ to $2000$.
In comparison, the model trained with GA-CTC loss predicts words accurately.
This matches our previous conclusion that GA loss is essential for handling large bias phrase lists.

\section{Conclusion}\label{sec: conclusion}

In this paper, experiments show that the proposed Guided Attention loss improves the effectiveness and robustness of Contextual Adapter without introducing additional parameters.
Compared to the alternative formulation, the proposed method is easier to implement and can achieve the same level of effectiveness when in practice.
Moreover, the proposed loss is not mutually exclusive with other methods which can potentially further improve performance, such as~\cite{AdaptiveCB,GatedCA,CACPP}.
Future experiments can be conducted to investigate the effect of different training bias list generation, such as one proposed in a recent research~\cite{wu23e_interspeech}.
Additionally, early-stage phrase filtering~\cite{wu23e_interspeech,yang23o_interspeech} might be an alternative method for handling large bias lists.

\vfill\pagebreak

\bibliographystyle{IEEEbib}
{\footnotesize \bibliography{main}}

\end{document}